**Behavioral Persistence and Incomplete Functional Transfer of Co-evolved Communication in Evolutionary Robotics**

Fernando Montes-Gonzalez

Instituto de Investigaciones en Inteligencia Artificial, Universidad Veracruzana, Xalapa, Veracruz, Mexico

ORCID: 0000-0002-8024-3023

Correspondence: fmontes@uv.mx

**Abstract**

This work evaluates the direct transfer of a co-evolved communication protocol from a 2D simulation to a 3D physical environment, without retraining the network weights. Two e-puck-type robots, controlled by a GRU network with residual connection, were evaluated in a food-seeking task with social signaling. The sensory and motor translation layer required three corrections for stable physical operation, including the calibration of a hunger term based on a measurable asymmetry in the trained residual weights. Even with these corrections, the transfer was partial and asymmetric: one agent reached the food source in one of thirty tested seeds, while the other did not reach it in any. Task success was measured by both agents reaching the food area. An additional experiment incorporating explicit directional information in the social channel produced observable changes in the trajectory of the receiving agent and improvements in several specific cases. However, these improvements were not enough to allow the second agent to reach the food source, suggesting that the limitation may not be explained solely by signal translation, but also by the ability to navigate under the new physical constraints. The results suggest that successful transfer of emergent communication may depend not only on preserving the signaling process itself, but also on preserving the ecological and navigational conditions under which the protocol evolved.

## 1. Introduction

Coevolved communication arises in multi-agent systems when two or more controllers, without an explicit programmed design, learn to coordinate their behavior using a shared signal. This design develops within a specific simulation environment with its own unique physical dynamics. This dependency raises the question: how well does a coevolved protocol survive when moved to a different physical environment without retraining the weights that generated it?

Most work on sim-to-real transfer in robotics has focused on individual controllers, tasked with a single navigation or manipulation function. Transferring a social communication protocol between two agents is a different problem. The signal must not only be translated into a new sensory and motor space, but it must also remain useful to the receiving agent under physical constraints that did not exist during the original learning process. The leap from a 2D simulation to a 3D physical simulation introduces such changes: friction, inertia, and collision dynamics that have no direct equivalent in the original environment.

This work directly evaluates this transfer, without retraining, in a food-seeking task with two e-puck-like robots. The contribution is not only the observation that the transfer is partial, but also identifying a structural asymmetry through a direct analysis of the residual weights in the neural network. It shows that one of the two agents, when presented with a variable that could be interpreted as *"fear"*, assigns it disproportionately greater weight than the term *"hunger"*. This asymmetry, tolerable in the original environment, may become a limiting factor under the new physical dynamics. Based on this, the hypothesis of "*implicit spatial coding"* is proposed, which stems from the idea that a co-evolved protocol does not carry all its useful information in the signal itself, but also in the physical context where it was learned. Furthermore, we conducted an additional experiment, injecting explicit directional information into the social channel, to directly test this idea.

In this work, we hypothesize that the partial transfer observed is not primarily due to a loss of information in the communication channel itself, but rather to the receiving agent's inability to utilize that information under a different physical dynamic. According to this hypothesis, increasing the signal's information content should not be sufficient to restore the agent's performance until its navigation capabilities are readjusted.

The remainder of the article is organized as follows. Section 2 introduces the related work. Section 3 describes the 3D simulation environment, the robot model, the controller architecture, and the adaptations required for physically stable transfer. Section 4 presents the results of the baseline transfer, the course correction experiment, and the social channel analysis. Section 5 discusses these results in light of the asymmetry measured in the residual weights and the concept of implicit spatial coding. Section 6 outlines the limitations regarding the interpretation of our results. Finally, Section 7 summarizes the conclusions and proposes directions for future work.

## 2. Related Work

The idea that intelligent behavior arises from the direct interaction between an agent and its physical environment, with subsumed behaviors, was brought to light by Brooks's work [1]. Pfeifer and Bongard [2] extend this idea by showing that the morphology and physical properties of the body both constrain and enable the forms of cognition that an agent can develop. This perspective is relevant to the present work because the communication protocol evaluated here did not co-evolve abstractly, but rather was embedded in a particular body and physical environment. Its transfer to a different body and environment provides an opportunity to examine the extent of this dependence between cognition and physical substrate.

Emergent communication in evolutionary robotics has been studied primarily in agent populations that co-evolve a signal and a response to that signal simultaneously. Marocco, Cangelosi, and Nolfi [3] showed that this type of protocol can emerge without explicit design in categorization tasks through physical interaction, and that it is strongly dependent on the sensory context in which it is trained. Mirolli and Nolfi [4] reviewed various evolutionary conditions that favor or hinder the emergence of communication between agents, including the role of the genetic relationship between sender and receiver. In a more recent study, Aldana-Franco, Montes-González, and Nolfi [5] analyzed the evolutionary utility of emergent communication systems and their relationship to signal complexity in foraging tasks, using the FARSA simulator and the Marxbot robot. The present work differs from these precedents in that it does not evaluate the emergence of the protocol itself, but its subsequent transfer to a physical environment different from that in which it was co-evolved.

In multi-agent reinforcement learning, emergent communication faces a similar problem of dependence on the context in which it was learned. Mordatch and Abbeel [6] show that agent populations can develop a compositional language that emerges within their training environment, but whose meaning remains tied to that environment. Lazaridou and Baroni [7], in their review of the area, point out that many emergent communication protocols remain strongly tied to the context in which they were learned. More recently, Li et al. [8] show that it is possible to achieve some generalization of a learned communication protocol to teammates, provided that the protocol relies on a shared representation, such as natural language. The present work evaluates a different case, where such an explicit shared representation does not exist: the protocol was co-evolved solely based on its usefulness for the task, without any additional mechanism that constrains its interpretation outside the training environment.

The transfer of controllers from simulation to a physical environment has been extensively studied under the term *"reality gap"*. Koos, Mouret, and Doncieux [9] propose a multi-objective approach that simultaneously optimizes simulation performance and controller transferability, based on the observation that the most efficient solutions in simulation often exploit imperfections or simplifications in the simulated model that are not present in the physical world. Salvato, Fenu, Medvet, and Pellegrino [10] offer a broader review of this problem in the context of reinforcement learning, documenting strategies such as domain randomization and co-simulation. Most of these works address the transfer of a single controller responsible for a single task. The present work addresses a different case: the transfer of a social protocol between two agents, where the signal must remain functionally useful to the receiving agent under new sensory, motor, and physical conditions.

The basal ganglia-inspired layer used in this work is conceptually motivated by Prescott, Montes-González, Gurney, Humphries, and Redgrave [11], who studied how simulated dopamine modulation affects action selection in a bioinspired basal ganglia model embedded in a foraging task. Furthermore, in their work, the authors simulated fear and hunger motivations to modulate foraging behavior. The present work does not implement a basal ganglia model like this, but it employs an action selection mechanism to reduce oscillation between competing actions during transfer to the three-dimensional simulator.

The literature on the reality gap traditionally addresses this problem as a question of equivalence between a simulator and the actual physical robot. The variable that is sought to be controlled is how well the simulator approximates real physical conditions, regardless of whether that simulator operates in a two- or three-dimensional space. This paper focuses on a distinct case: the transfer between two physical simulators that differ in dimensionality, one restricted to a plane and the other incorporating 3D contact and friction dynamics, prior to any transfer to real hardware. This jump in physical dimensionality is studied here as a particular form of transfer gap related to the broader simulation-reality transfer problem, applied specifically to the case of a co-evolved social communication protocol between two agents.

## 3. Materials and Methods

### 3.1 Original 2D evolution environment

The controllers used in this work were previously obtained through a coevolution process carried out in a two-dimensional environment developed in Pygame [12]. In this environment (see Figure 1), two mobile agents with differential locomotion had to locate a food source within an arena with static obstacles generated in random positions. In addition to local perception of the environment using virtual sensors, the agents could exchange social signals as part of the coordination process.

The task considered during evolution corresponds to the general cooperative search problem used in the transfer experiments described in this article. The original environment used a simplified two-dimensional representation, whereas the evaluation performed in this work was carried out in a three-dimensional, physics-based environment. The weights of the evolved controllers were transferred without retraining to the new environment, allowing us to assess the extent to which previously acquired behaviors and communication strategies were preserved.

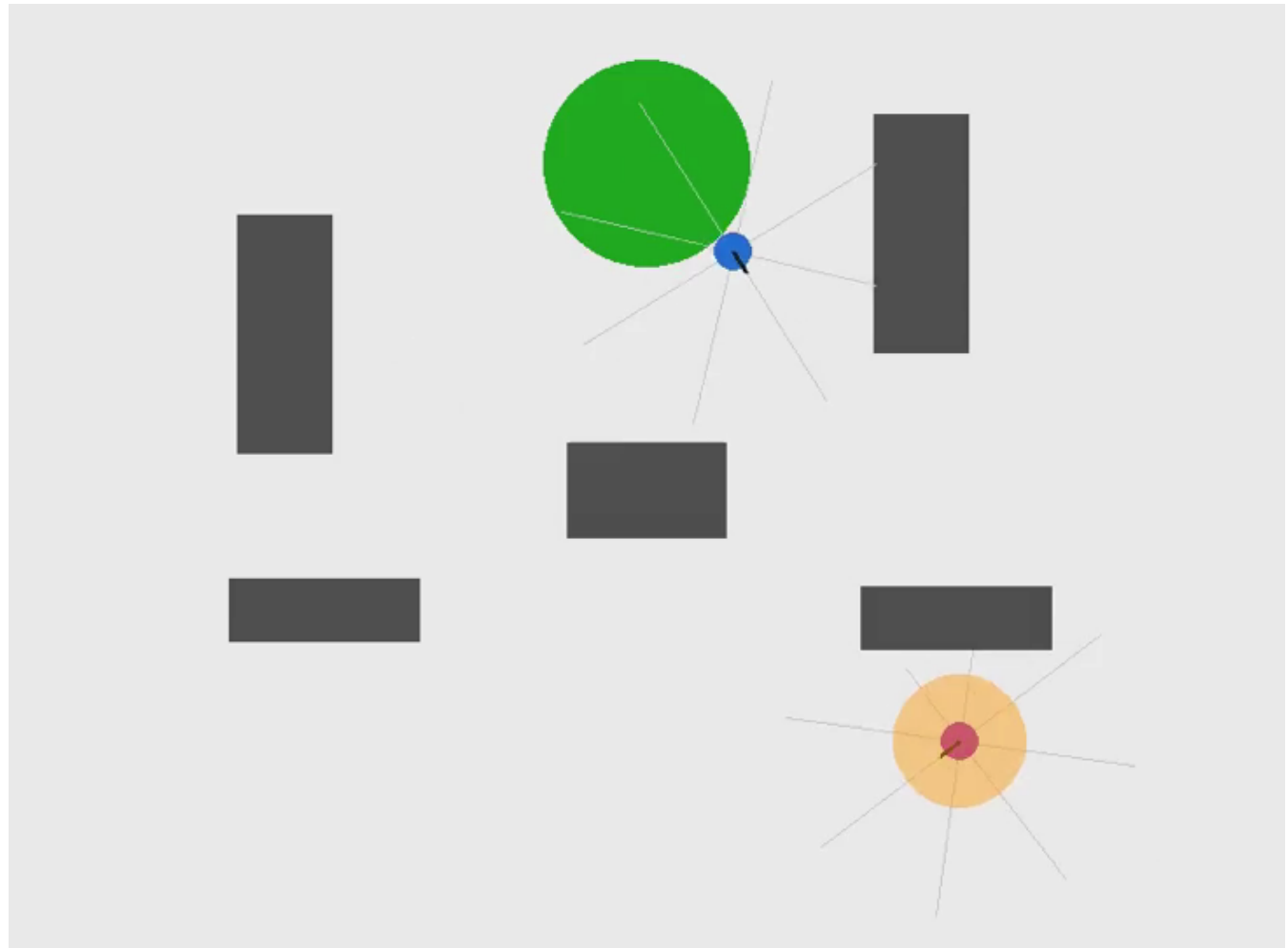

**Figure 1.** Original two-dimensional environment implemented in Pygame. The arena contains static obstacles distributed in random positions, a food source (green circle), and two mobile agents equipped with proximity sensors. The yellow halo around one of the agents represents the emission of a social signal.

### 3.2 Simulation environment with 3D physics engine

The three-dimensional transfer environment was implemented in PyBullet [13]. The transfer environment consists of a 1.5-meter square arena, constructed from static geometry of walls and random obstacles loaded from a JSON file that defines the environment. Gravity was set to -9.81 m/s². The physics engine updates at a time step of 0.00417 seconds. Two differentially driven, e-puck-type robots are loaded from a shared URDF model. At the start of the episode, these are placed within the arena in valid positions with random orientations that do not overlap with existing obstacles. The robots' initial positions are generated maintaining a minimum separation of 0.075 m from the walls.

### 3.3 Robot and sensor model

Each robot has a ring of eight proximity sensors spaced 45 degrees around its body (see Figure 2). Readings are obtained by ray tracing from the robot's center, with a maximum range of 0.20 meters per ray. A reading of 1.0 indicates an obstacle on the robot's surface, and 0.0 indicates the absence of a nearby obstacle. Rays intersecting the robot's own body are excluded from detection. The robot's wheel radius is 0.02 meters, and its maximum linear speed is 0.15 m/s.

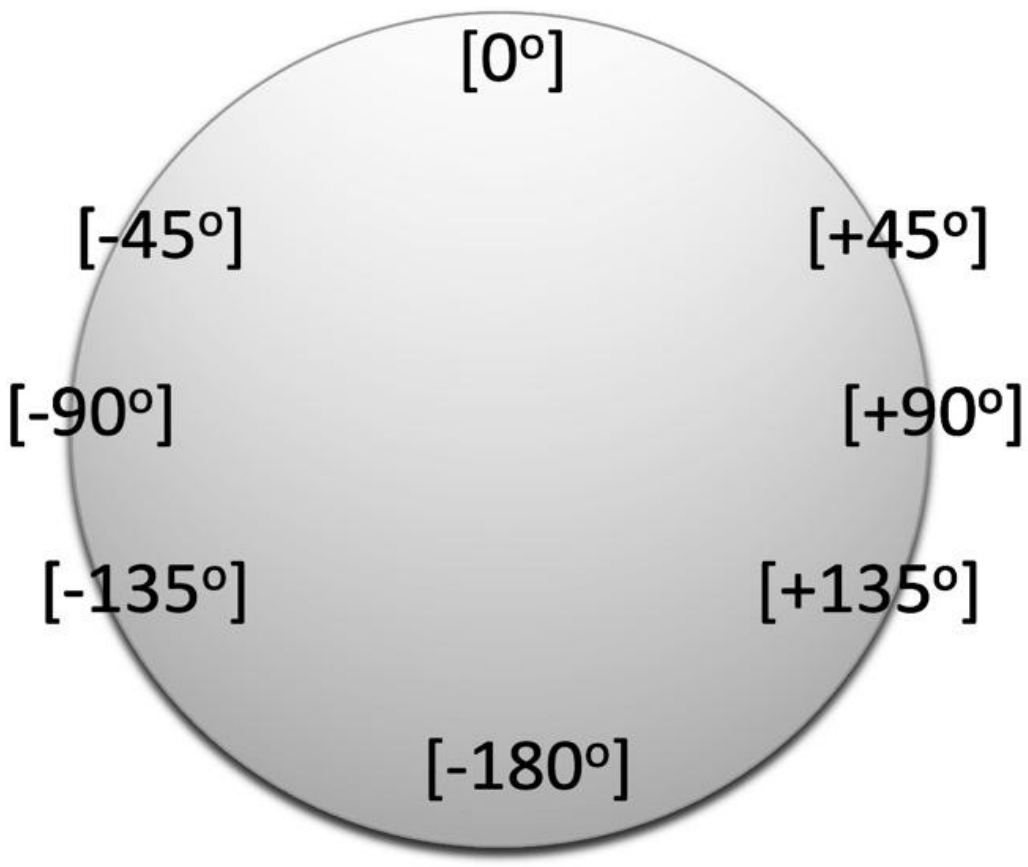

**Figure 2.** Distribution of the robot's eight proximity sensors. The sensors are evenly spaced at 45° intervals around the body, providing 360° coverage.

### 3.4 Controller Architecture

Each robot is controlled by a recurrent-linked GRU network (RES-GRU), which has eleven input units and sixteen hidden units. The input vector includes eight proximity readings, the relative bearing to the food source, a hunger term calculated as the exponential decay of the distance to the food, and a binary flag indicating whether the robot is currently stopped. The network output is a discrete action, selected by argmax from five possible actions: *forward*, *turn left*, *turn right*, *reverse*, and a *signaling action*. In addition to the recurrent path, there is a direct residual link that maps the input vector to the output logits. A basal ganglia-inspired layer temporarily replaces the network output with a fixed escape action for several simulation frames when the robot is detected as stuck, in order to reduce oscillation between actions competing to control the motors. The controller weights for both agents come from a previous coevolution process in a 2D simulation environment, and were not modified during the experiments of this work, except as indicated in the following section.

### 3.5 Translation layer adaptations for 3D transfer

The direct transfer of the 2D trained weights to the 3D physical environment required three adjustments to the layer that translates the controller output to the physical motor. First, the mapping of turn actions to wheel speeds was corrected. In the original 2D implementation, turn actions produced pure rotation about the robot's axis, with no forward translation. An earlier 3D implementation applied differential speeds while maintaining a constant forward component on both wheels, causing the robot to continue moving forward while attempting to turn to avoid an obstacle. As a result, it tended to remain in contact with walls and slide along them instead of separating after the evasive maneuver. To approximate the rotational behavior of the original 2D controller, the residual forward momentum present during turn maneuvers was reduced. Second, the implementation of the filter regulating the injection of the social signal into the front sensor input was revised and corrected.

Third, the hunger term in input x[9] was recalibrated for one of the two agents. This recalibration was based on a direct analysis of the residual weights. Agent B's residual connection exhibits a greater relative importance to proximity cues associated with obstacle avoidance, which may be interpreted analogously as a fear component. The relationship between this component and the hunger term is approximately 1.78 times greater in Agent B than in Agent A. This asymmetry was estimated by calculating the norm of the corresponding rows of the residual weight matrix for each input. To compensate for this without altering the evolved weights, the divisor of the exponential decay that determines x[9] was adjusted instead of directly scaling its amplitude, thus preserving the range [0, 1] observed during 2D training. The decay value was incrementally evaluated with values between 1.0 and 1.25 using the same runs with fixed seeds. The value of 1.25 was retained for Agent B in the experiments reported in this work.

3.6 Bearing Fix Experiment

An additional, separate modification to the translation layer, not included in the baseline configuration described in the previous section, was evaluated. This modification replaced the value of x[8], which normally encodes the relative direction toward the food source, with the relative direction toward the agent emitting the social signal. The rewrite was performed only when the other agent activated the social signal and the hunger term of the receiving agent remained below 0.6. The aim was to test whether the absence of explicit directional encoding in the social channel could contribute to the reduced performance of Agent B. This modification was evaluated independently using the same thirty fixed seeds, and was not incorporated into the baseline configuration used in the other experiments reported in this work.

3.7 Experimental protocol

Each evaluation run used a fixed seed to control the initial position and orientation of the agents, allowing comparison with control conditions. The controller makes a decision every 15 physical simulation steps, equivalent to approximately 62.5 milliseconds of simulated time. An episode ends when both agents reach the food source, defined as a Euclidean distance of less than 0.08 meters, or upon reaching a maximum simulation step limit.

In each simulation step, the following were recorded for each agent: *eight proximity readings*, *relative heading toward the food source*, *hunger term*, *stuck flag*, *robot position*, and *orientation*. Two metrics were calculated from these records. The sign-change rate (**saw_flip_rate**) measures how frequently the hunger term alternates between significant increases and decreases between consecutive steps, ignoring minor variations attributable to noise. This metric captures short-term oscillatory patterns associated with sawtooth-like trajectories in which the agent repeatedly alternates between moving toward and away from the food source. The cycle rate (**saw_cycle_rate**) counts complete oscillation cycles, defined as two consecutive sign changes (e.g., positive-negative-positive or negative-positive-negative) that return the trend to the initial state. Like saw_flip_rate, this metric is expressed as a relative frequency throughout the episode.

In addition to the directly recorded variables, the **magnet_max** metric was calculated, defined as the maximum value of the hunger term x[9] during a run. This term is calculated as *x[9] = exp(-dist/hunger_decay)*, where *dist* is the Euclidean distance between the robot and the food source. Since x[9] increases monotonically and exponentially as the distance to the food decreases, magnet_max provides a summary measure of the maximum proximity achieved by an agent during a run. Values close to 1 indicate proximity to the food, while low values indicate that the agent remained far from the target.

3.8 World Configuration

The physical simulation environment consists of a square arena enclosed by red walls (see Figure 3). Inside, there is a food area and five rectangular gray obstacles. The food source is represented by a yellow circle on the simulation plane and a yellow sphere raised above it to improve its visibility during the experiments. The obstacles are distributed in random positions at the start of each simulation, generating different configurations that alter the navigation paths available to the robots.

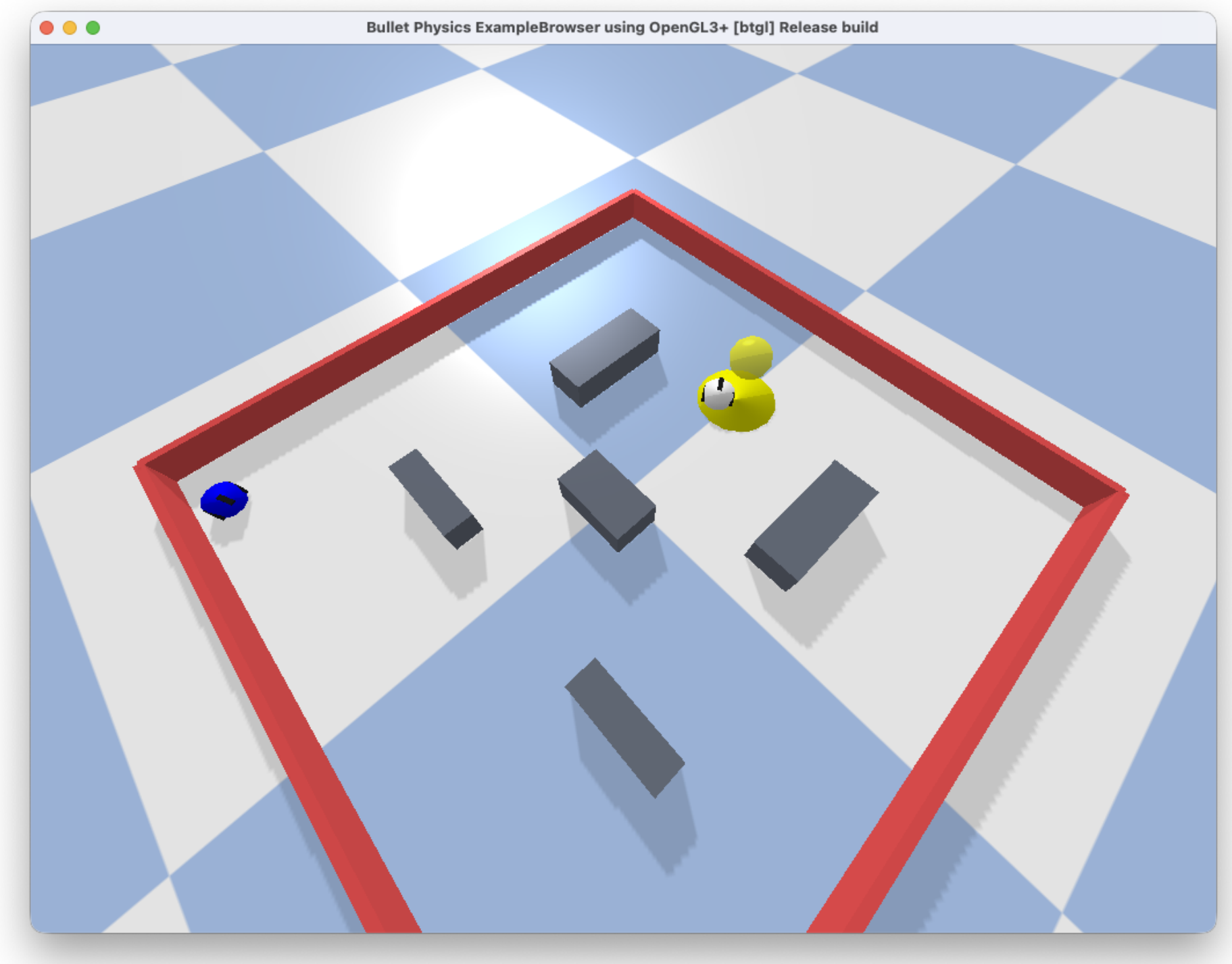

**Figure 3.** Experimental environment implemented in PyBullet. The arena contains five rectangular obstacles and a food area represented in yellow. The robots alternate between their identity colors (white and gray) and state colors used to visualize signaling states (**blue**: signal reception; **magenta**: signal transmission, not visible in this figure).

The agents are two e-puck-like robots controlled by recurrent neural networks. To facilitate visual analysis of their behavior, the robots use a color scheme associated with different internal states. Under normal conditions, Robot A is represented in white and Robot B in gray. When a robot is physically blocked (stalled), its color changes to red. The blue color on Robot B indicates that the agent is receiving a social signal from the other robot, while the magenta color on Robot A indicates that it is actively emitting a social signal. Additionally, when the distance to the food source is less than 0.4 m, both robots can be displayed in yellow to indicate proximity to the target resource. This visual coding scheme allows for the rapid identification of signaling states, physical blockages, and successful approaches to the food source during transfer experiments.

## 4. Results

### 4.1 Baseline Transfer

The results reported in Table 1 show the performance of both agents across 30 evaluation seeds under the baseline transfer condition. Agent A achieved a mean *magnet_max* value of 0.820, with a standard deviation of 0.106, ranging from 0.567 to 0.968. Agent B achieved a mean *magnet_max* value of 0.583, with a standard deviation of 0.110, ranging from 0.422 to 0.852. Using a success threshold of less than 0.08 m from the food source, Agent A reached the food source on 1 of the 30 seeds. Agent B did not reach the food source on any seeds.

| Seed | A magnet_max | B magnet_max | Seed | A magnet_max | B magnet_max |
|---|---|---|---|---|---|
| s1 | 0.907 | 0.658 | s16 | 0.866 | 0.659 |
| s2 | 0.646 | 0.655 | s17 | 0.968 * | 0.476 |
| s3 | 0.867 | 0.658 | s18 | 0.859 | 0.628 |
| s4 | 0.895 | 0.448 | s19 | 0.686 | 0.517 |
| s5 | 0.923 | 0.621 | s20 | 0.568 | 0.505 |
| s6 | 0.782 | 0.475 | s21 | 0.677 | 0.549 |
| s7 | 0.823 | 0.582 | s22 | 0.922 | 0.493 |
| s8 | 0.866 | 0.468 | s23 | 0.580 | 0.607 |
| s9 | 0.836 | 0.519 | s24 | 0.918 | 0.480 |
| s10 | 0.917 | 0.471 | s25 | 0.920 | 0.852 |
| s11 | 0.809 | 0.637 | s26 | 0.811 | 0.422 |
| s12 | 0.920 | 0.441 | s27 | 0.800 | 0.745 |
| s13 | 0.873 | 0.650 | s28 | 0.920 | 0.626 |
| s14 | 0.740 | 0.495 | s29 | 0.836 | 0.681 |
| s15 | 0.691 | 0.656 | s30 | 0.763 | 0.819 |

**Table 1**. *magnet_max* values per seed for both agents under the base transfer condition, across 30 seeds. The asterisk (*****) indicates the only seed in which an agent reached the food source.

The mean *saw_flip_rate* was 0.041, with a standard deviation of 0.017, for Agent A and 0.093, with a standard deviation of 0.116, for Agent B. The mean *saw_cycle_rate* was 0.020 for Agent A and 0.043 for Agent B. On average, Agent B exhibited higher swing rates than Agent A, although with substantially greater variability among the seeds. This is consistent with more frequent alternations between approach and retreat relative to Agent A.

4.2 Orientation Correction Experiment (Bearing Fix)

The bearing fix experiment evaluated the hypothesis that incorporating explicit directional information into the social channel could improve Agent B's performance after transfer to the three-dimensional environment. During social signaling events, x[8] was replaced by the relative bearing toward the sending agent. Under normal conditions, x[8] represents the relative bearing toward the food source. The replacement occurred only when the receiving agent's hunger level remained below a predefined threshold. This modification was applied exclusively to the translation layer between the controller and the physical environment, leaving the architecture and weights of the evolved controllers unchanged.

Across the same 30 evaluation seeds (see Table 2), Agent B still failed to complete the task successfully, even under the bearing fix condition. However, the fix modified the *magnet_max* values in 11 of the 30 seeds evaluated (see columns “B base” and “B bearing”). Of these 11 seeds, *magnet_max* increased in 7 compared to the baseline condition, with increases ranging from 0.036 to 0.322. The largest improvement was observed in seed s9. In the remaining 4 seeds, *magnet_max* decreased slightly, with a maximum reduction of 0.037 in seed s10. In the other 19 seeds, the values remained identical to those obtained in the baseline runs. These results indicate that the bearing fix was able to alter *magnet_max* in some specific cases. However, this effect was not consistent across seeds, nor did it result in both robots reaching the food. For Agent A (see columns “A base” and “A bearing”), the correction's influence was also limited: *magnet_max* decreased in 8 seeds, increased in 6, and remained unchanged in the remaining 16, with no evidence of a consistent performance improvement.

| Seed | A base | A bearing | B base | B bearing | Seed | A base | A bearing | B base | B bearing |
|---|---|---|---|---|---|---|---|---|---|
| s1 | 0.907 | 0.880 | 0.658 | 0.658 | s16 | 0.866 | 0.916 | 0.659 | 0.647 |
| s2 | 0.646 | 0.616 | 0.655 | 0.764 | s17 | 0.968 | 0.968 | 0.476 | 0.472 |
| s3 | 0.867 | 0.867 | 0.658 | 0.658 | s18 | 0.859 | 0.627 | 0.628 | 0.628 |
| s4 | 0.895 | 0.639 | 0.448 | 0.546 | s19 | 0.686 | 0.610 | 0.517 | 0.517 |
| s5 | 0.923 | 0.923 | 0.621 | 0.657 | s20 | 0.568 | 0.568 | 0.505 | 0.484 |
| s6 | 0.782 | 0.815 | 0.475 | 0.475 | s21 | 0.677 | 0.678 | 0.549 | 0.549 |
| s7 | 0.823 | 0.921 | 0.582 | 0.582 | s22 | 0.922 | 0.922 | 0.493 | 0.493 |
| s8 | 0.866 | 0.866 | 0.468 | 0.468 | s23 | 0.580 | 0.580 | 0.607 | 0.607 |
| s9 | 0.836 | 0.809 | 0.519 | 0.841 | s24 | 0.918 | 0.918 | 0.480 | 0.466 |
| s10 | 0.917 | 0.878 | 0.471 | 0.434 | s25 | 0.920 | 0.920 | 0.852 | 0.852 |
| s11 | 0.809 | 0.809 | 0.637 | 0.637 | s26 | 0.811 | 0.842 | 0.422 | 0.422 |
| s12 | 0.920 | 0.920 | 0.441 | 0.623 | s27 | 0.800 | 0.800 | 0.745 | 0.745 |
| s13 | 0.873 | 0.873 | 0.650 | 0.650 | s28 | 0.920 | 0.920 | 0.626 | 0.626 |
| s14 | 0.740 | 0.922 | 0.495 | 0.495 | s29 | 0.836 | 0.836 | 0.681 | 0.681 |
| s15 | 0.691 | 0.893 | 0.656 | 0.934 | s30 | 0.763 | 0.667 | 0.819 | 0.936 |

**Table 2.** Comparison of *magnet_max* values between the baseline and bearing fix runs for both agents across 30 seeds.

These results provide relevant information in two aspects. First, they suggest that signal translation alone is unlikely to explain the observed transfer limitations. Although the bearing fix introduced explicit directional information and led to improvements in several seeds, Agent B still failed to complete the foraging task in any of the 30 evaluations. Second, it indicates that this correction modifies the receiver's behavior and can improve signal tracking in some seeds. However, these changes were insufficient to restore successful task transfer. The communication channel appears to retain its behavioral relevance, although its effects on task performance remain asymmetric between agents.

### 4.3 Social Channel Analysis

The results of the bearing fix experiment suggest that incorporating explicit directional information can modify, in some seeds, the behavior reflected by *magnet_max*. However, these modifications were not consistent across runs, nor did they allow Agent B to cooperate with Agent A so that both could reach the food. Overall, these results are consistent with the possibility that transferring to the three-dimensional environment involves limitations that are unlikely to be resolved solely by adding explicit directional information to the social channel. This issue is discussed in more detail in the next section.

## 5. Discussion

The results presented in this article suggest that the direct transfer of an emergent co-evolved communication protocol from a 2D environment to a 3D physical simulation shows partial and asymmetric performance. By exploring the environment, Agent A is able to reach the food source in one of the thirty seeds. Agent B, however, fails to reach the food source on any occasion. The bearing fix experiment suggests that the bottleneck is not solely explained by signal translation. Although the additional directional information produced improvements in several cases, it was never sufficient to allow Agent B to successfully complete the task under the new conditions. This result is consistent with our initial hypothesis that the observed transfer limitations may originate from the receiver's inability to effectively utilize the transmitted information following the change in physical dynamics, rather than from the signal containing insufficient information.

One possible explanation is that the co-evolved protocol was shaped within a 2D environment where physical proximity and signal activation were closely linked. The results are consistent with the possibility that Agent B associated the social signal with a spatial configuration, where the presence of Agent A near the food source guides its trajectory. In 3D, this link appears to be altered not only by changes in the spatial context but also by an imbalance in Agent B's evolved weights. Agent B's residual connection exhibits a greater relative importance to proximity signals associated with obstacle avoidance, which may be interpreted as fear, than to the hunger signal. In particular, the relationship between fear and hunger is approximately 1.78 times stronger in Agent B than in Agent A. This imbalance was tolerable in the two-dimensional environment but may be a limiting factor in the physical dynamics of the transfer within a three-dimensional environment. The observed behavioral responses suggest that certain elements of the signaling process survive the transfer. However, this was insufficient to produce successful navigation behavior by Agent B under the evaluated conditions.

These results are consistent with the hypothesis of implicit spatial coding in emergent communication; it is possible that the informational content of the protocol does not reside in the signal itself, but also depends on the physical context in which it evolved. This is analogous to the simulation-to-real transfer problem in robotics, where controllers that work well in simulation may fail when physically implemented; this may occur even when the control logic itself remains unchanged, but because the sensorimotor correspondence available during training was more stable than that available after the transfer.

Before the bearing fix experiment, an independent calibration of the residual inputs was attempted. Analysis of the residual weights revealed the previously described asymmetry between the fear and hunger-related components in Agent B. Therefore, the hunger term was recalibrated using the decay constant applied to x[9], maintaining its value within the range observed in the 2D training, which was [0, 1]. For recalibration, a range of values from 1.0 (the baseline) to 1.25 was used, employing repeated runs in the same order with different seeds. The results depended on the seed used.

The baseline results presented in Table 1 already reflect the best calibration with a decay value of 1.25 for Agent B. Even with this calibrated baseline configuration, Agent B failed to reach the food source in any of the thirty seeds evaluated. This suggests that adjusting the input at the residual level is not sufficient to restore successful transfer performance, similar to the bearing fix experiment.

The bearing fix experiment is consistent with this interpretation. This fix increased the information available to the receiving agent (B) by providing an explicit directional signal that was not present in the original communication channel. Despite this increase in available information, the observed increases in magnet_max for several seeds did not translate into successful orientation to the food. Incorporating explicit directional information into the signal channel was insufficient to restore Agent B's performance. This suggests that the problem may not be explained solely by the information available in the signal. The results could reflect limitations in navigation toward the food source within the three-dimensional environment, limiting the effective use of available social information.

The findings above suggest implications for the design of multi-agent evolutionary systems capable of extending to different simulation environments or potentially to physical robots. In particular, the results suggest that the transferability of a co-evolved protocol should not be automatically assumed when the spatial and sensorimotor conditions of the environment change. In future work, one possible approach would be to retrain only the receiving agent while keeping the sender's weights fixed before performing a partial re-evolution. This could help restore the functional reciprocity of the protocol without requiring an entirely new co-evolution process.

Another alternative would be to employ world models as an intermediate layer. If the GRU-based controller already encodes an internal representation of the environment acquired during 2D evolution, this representation could potentially be adapted to a three-dimensional environment through latent space refinement. In this way, agents could practice coordination within a simplified model of the environment before interacting with the full physics engine, potentially reducing the computational cost associated with physics simulation. This approach is consistent with the concept of *“latent imagination”*, in which agents learn behaviors from a world model learned through reinforcement learning [14]. It is also consistent with the *“World Models”* framework [15] and represents a possible extension of the architecture used in this work and a plausible direction for future research.

## 6. Limitations

This study has several limitations that should be considered when interpreting the results. The evaluation was conducted using thirty fixed seeds. This number expands the experimental coverage compared to the preliminary tests and allows for more robust comparisons between conditions. However, the reported results should not be interpreted as a comprehensive statistical characterization of all possible environmental configurations. Given the exploratory nature of this evaluation, the analysis focused on descriptive metrics and direct comparisons between experimental conditions.

The bearing fix experiment did not include a detailed record of social signal activation throughout each episode. Since the objective of this evaluation was to determine whether the proposed correction affected the final performance of the receiving agent, the analysis focused on the outcome metrics recorded for each seed. Consequently, a comprehensive characterization of the signal's temporal dynamics during the experiment was not performed, since it was outside the scope of the evaluation.

This study does not include a direct comparison with the performance of the same controllers in their original 2D environment. Such a comparison would have allowed for a more precise quantification of the magnitude of the degradation associated with the transfer. It was omitted because the corresponding 2D evaluation is currently under review. Additionally, the findings of this study correspond to a single pair of coevolved controllers and to the foraging task of both agents in a single food area. It was not assessed whether the asymmetry measured in the residual weights can be generalized reliably to other pairs of controllers or to other coevolved tasks. Likewise, it remains unclear whether the implicit spatial coding mechanism proposed in this study can be generalized beyond the specific controllers analyzed here. Consequently, the proposed implicit spatial coding mechanism should be interpreted as a working hypothesis consistent with the observed results, rather than as a proven general property of co-evolved communication systems.

## 7. Conclusions

This work evaluated the direct transfer of an emergent co-evolved communication protocol from a 2D simulation to a 3D physical simulation environment. A RES-GRU controller with a reactive layer inspired by the basal ganglia was used as the robot's control module, without retraining the network weights.

The main findings can be summarized as follows. First, Agent A retained partial foraging ability and reached the food source in one of the thirty seeds tested. Second, Agent B did not reach the food source in any of the thirty seeds, indicating an asymmetric transfer outcome between agents. Third, recalibrating the hunger term by adjusting the residual pathway was insufficient to restore full functional transfer, as Agent B failed to reach the food in any of the tests. Fourth, a targeted experiment incorporating explicit directional information into the communication channel produced observable changes in the receiving agent's behavior and performance improvements in several specific cases. However, these improvements were never sufficient to allow Agent B to reach the food source, suggesting that the problem may not be explained solely by a lack of directional information in the signal.

These results suggest that the effectiveness of the transferred communication protocol depended on environmental conditions that were altered during the transition from 2D to 3D simulation. Behavioral evidence suggests that some aspects of the signaling process are transferred, while the environmental conditions associated with that signaling may not be fully preserved for maintaining effective coordination. The bearing fix experiment suggests that introducing explicit directional information into the signal channel is insufficient to restore the behavior obtained in 2D. Achieving full functional transfer may require the receiving agent to re-evolve its navigational capabilities in a 3D environment.

This work highlights a distinction between the transfer of an evolved signaling mechanism and the transfer of the coordinated behavior supported by that mechanism. These findings suggest partial re-evolution and latent-space world models as potential strategies for achieving robust transfer of coevolved coordination between two-dimensional and three-dimensional simulators.

**Declarations**

Funding

This work was supported by the National System of Researchers (SNII, Mexico). Fernando Montes-González is a member of the National System of Researchers (SNII), Researcher ID 30026.

Competing Interests

The author has no competing interests to declare.

Data Availability

The source code, simulation environments, and data analysis scripts associated with this study are permanently archived in Zenodo at https://doi.org/10.5281/zenodo.21731483 (Version v1.0.0). Ongoing development and updates are available at the GitHub repository: https://github.com/ferdiex/essim

Author Contributions

Fernando Montes-González conceived the study, developed the methodology, performed the experiments, analyzed the results, and wrote the manuscript.